\documentclass{llncs}

\usepackage{url}            
\usepackage{booktabs}       
\usepackage{amsmath}
\usepackage{hyperref}
\usepackage{amssymb}
\usepackage{mathtools}
\usepackage{graphicx}

\DeclareMathOperator*{\argmax}{arg\,max}

\title{Learning Multimodal Transition Dynamics for Model-Based Reinforcement Learning}
\author{Thomas M. Moerland, Joost Broekens, and Catholijn M. Jonker}
\institute{Department of Computer Science \\ Delft University of Technology, The Netherlands \\ \email{\{T.M.Moerland,D.J.Broekens,C.M.Jonker\}@tudelft.nl}}

\begin{document}

\maketitle

\begin{abstract}
In this paper we study how to learn stochastic, multimodal transition dynamics in reinforcement learning (RL) tasks. We focus on evaluating transition function estimation, while we defer planning over this model to future work. Stochasticity is a fundamental property of many task environments. However, discriminative function approximators have difficulty estimating multimodal stochasticity. In contrast, deep generative models do capture complex high-dimensional outcome distributions. First we discuss why, amongst such models, conditional variational inference (VI) is theoretically most appealing for model-based RL. Subsequently, we compare different VI models on their ability to learn complex stochasticity on simulated functions, as well as on a typical RL gridworld with multimodal dynamics. Results show VI successfully predicts multimodal outcomes, but also robustly ignores these for deterministic parts of the transition dynamics. In summary, we show a robust method to learn multimodal transitions using function approximation, which is a key preliminary for model-based RL in stochastic domains.
\end{abstract}

\section{Introduction} \label{introduction}
\begin{figure}[htb]
\centering
\includegraphics[scale = 0.35]{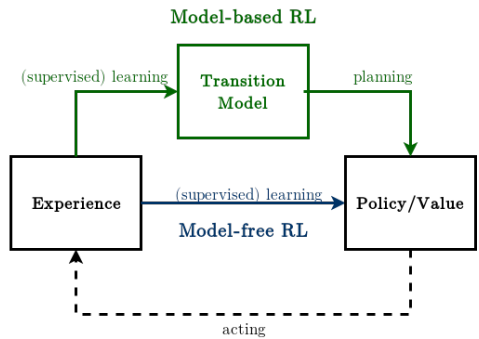}
\vspace{-0.3cm} 
\caption{Two types of reinforcement learning. {\it Model-free} RL (blue) directly learns a behavioural policy, while {\it model-based} RL (green) also attempts to learn the environment's transition dynamics. Learning this transition model allows the agent to predict future states and thereby allows the agent to {\it plan}.}
    \label{modelrl}
\vspace{-0.3cm}
\end{figure}

Reinforcement learning (RL) is a succesful learning paradigm for sequential decision making from data in agents and robots. A long standing debate in RL research has been whether to learn `model-free' or `model-based' (Figure \ref{modelrl}) \cite{sutton1991dyna}. Model-based RL has shown some important benefits, most noteworthy increased data efficiency \cite{deisenroth2011pilco}, the potential for targeted exploration \cite{stadie2015incentivizing}, and natural transfer between tasks when only the reward function changes. Model-based RL consists of two steps: 1) transition function estimation through supervised learning, and 2) (sample-based) planning over the learned model (Figure \ref{modelrl}, green arrows). Each step has a particular challenging aspect. For this work we focus on a key challenge of the first step, {\it stochasticity} in the transition dynamics, while we defer the second step, planning (under uncertainty), to future work (see Sec. \ref{future} as well).

Stochasticity is an inherent property of many environments, and increases in real-world settings due to sensor noise. Transition dynamics usually combine both deterministic aspects (such as the falling trajectory of an object due to gravity) and stochastic elements (such as the behaviour of another car on the road). Our goal is to learn to jointly predict these. Note that stochasticity has many forms, both homoscedastic versus heteroscedastic, and unimodal versus multimodal. In this work we specifically focus on multimodal stochasticity, as this should theoretically pose the largest challenge.

To learn such transition models, we require high-capacity function approximators that can predict next-state distributions of complex shape. This problem is not yet accurately solved by currently used methods in model-based RL, like tabular learning \cite{brafman2002r} (which does not scale to high-dimensions), linear function approximation \cite{atkeson1997locally} with Gaussian noise \cite{li2011knows}, random forests \cite{hester2012learning}, or deep feed-forward networks trained on mean-squared error (MSE) \cite{oh2015action} (See also Sec. \ref{challenges}). 

A potential solution to this problem are Deep Generative Models (DGM) \cite{goodfellow2016nips}, as they are non-linear function approximators that can learn complex outcome distributions and scale to high-dimensions. In Section \ref{challenges} we compare different DGM's on their theoretic appeal for stochastic model learning in the RL setting, and identify conditional Variational Inference (VI) as the most promising solution. 

The remainder of the paper then continues as follows. In Section 3, we formally describe conditional variational inference with different types of discrete and continuous latent variables. In Section 4 we empirically compare the different approaches on a simulated function and on a typical RL tasks. Our results show that VI is accurately able to discriminate deterministic from stochastic aspects of the transition dynamics. We also show how the RL agent manages to learn an accurate transition model while solving a task. Finally, Sections 5 and 6 connect our work to related literature and identify opportunities for future work, respectively. 

All code to reproduce the results in this paper is publicly available at \url{www.github.com/tmoer/multimodal_varinf}.

\begin{figure}[t]
\centering
\includegraphics[scale = 0.20]{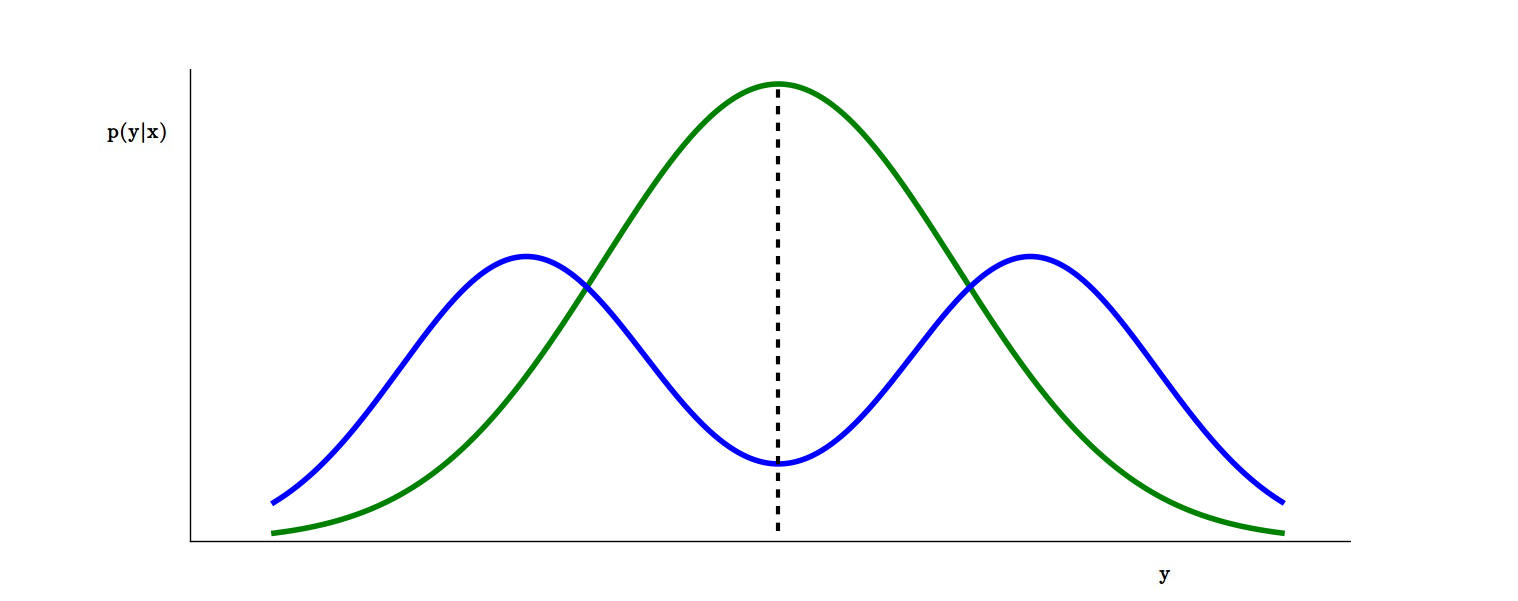}
\vspace{-0.4cm} 
\caption{Multimodal outcome distribution $p(y|x)$ (blue line) for a simple 1D observation space. Training on a mean-squared error will deterministically predict the conditional mean (dashed line), which implicitly assumes a uni-modal Gaussian outcome (green line).}
    \label{bimodal}
\vspace{-0.4cm}
\end{figure}

\section{Challenge of Multimodal Transitions} \label{challenges}
We will write $x \in \mathcal{X}$ for the current state and action, and $y \in \mathcal{Y}$ for the next state we want to predict. We are interested in models that can approximate distributions $p(y|x)$ with multiple local maxima (`modes'). A cardinal example of such a distribution is shown in Figure \ref{bimodal} (blue line). 

Discriminative function approximators, for example trained at mean-squared error (MSE) loss, fail at this task. The point prediction of the function approximator will be the conditional expectation of the outcome distribution (Fig. \ref{bimodal}, dashed line). Obviously, point estimate predictions will never be a good method to approximate a distribution, but they have actually been frequently applied in model-based RL work \cite{oh2015action}.\footnote{For example, Oh et al. \cite{oh2015action} shows MSE training does work well in high-dimensional, deterministic domains. However, for example inspecting their Ms. Pacman (a stochastic game) predictions at \url{https://youtu.be/cy96rtUdBuE}, we see that the predictions for the stochastic elements (ghosts) fail. The ghosts disappear when they reach a corridor junction, where they stochastically choose in which direction to continue. The feed-forward network predicts the conditional mean of these choices, which completely blurs the ghosts in a few frames.} Clearly, a unimodal outcome distribution will neither solve the multimodality problem (Fig. \ref{bimodal}).

A mixture of Gaussians per outcome dimension would neither solve the problem. Full covariance matrix Gaussians clearly do not scale to high-dimensional domains (such as \cite{mnih2015human}), while diagonal Gaussians would loose all covariance structure in the predictions. Moreover, mixture models are tedious to train. What we require are models that 1) flexibly approximate joint distributions of complex (multimodal) shape and 2) scale to high-dimensions.

\begin{figure}[tbp]
\centering
\includegraphics[scale = 0.28]{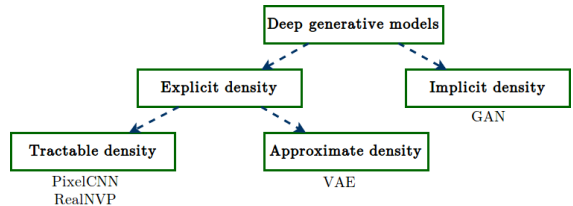}
\vspace{-0.3cm} 
\caption{Deep generative model taxonomy following \cite{goodfellow2016nips}. }
    \label{dgm}
\vspace{-0.4cm}
\end{figure}

We hypothesize the group of deep generative models (DGN) \cite{goodfellow2016nips} are a promising candidate, as they fulfill both requirements. For model-based RL, where we will use the models to sample (a lot of) traces, we additionally require that the model is 1) easy to sample from, and 2) ideally allows for planning at an abstract level. Following the DGN taxonomy by Goodfellow \cite{goodfellow2016nips} (Figure \ref{dgm}), we now compare DGN models on their theoretical appeal for transition function estimation. 

Implicit density models, like Generative Adverserial Networks (GAN) lack a clear probabilistic objective function, which was the focus of this work. Among the explicit density models, there are two categories. {\it Change of variable formula} models, like Real NVP \cite{dinh2016density}, have the drawback that the latent space dimension must equal the observation space. Fully visible belief nets like pixelCNN \cite{oord2016conditional}, which factorize the likelihood in an auto-regressive fashion, hold state-of-the-art likelihood results. However, they have the drawback that sampling is a sequential operation (e.g. pixel-by-pixel, which is computationally expensive), and they do not allow for latent level planning either. Therefore, most suitable for model-based RL seem approximate density models, most noteworthy the variational auto-encoder (VAE) \cite{kingma2013auto} framework. These models can estimate stochasticity at a latent level, allow for latent planning \cite{watter2015embed}, are easy to sample from, and have a clear probabilistic interpretation. In the next section, we will formally introduce this methodology in the conditional setting, where the generative process of $y$ is conditioned on other variables $x$.

\section{Conditional Variational Inference} \label{vae}
We will first introduce the conditional variational auto-encoder (CVAE) \cite{sohn2015learning}. Our goal is to learn a generative model of a (possibly multimodal) distribution $p(y|x)$. We assume the generative process is actually conditioned on latent variables $z$:
\vspace{-0.2cm}
\begin{equation}
p(y|x) = \int p(y|z,x) p(z|x) dz
\end{equation} 

Here $p(z|x)$ is the {\it prior} and $p(y|z,x)$ is the {\it generative model} or `decoder'. The stochastic latent variables $z$ provide the flexibility to predict more complex outcome distributions $y$. The posterior over $z$, $p(z|y,x)$ is intractable in most models of interest, for example deep non-linear neural networks. However, the parameters of this distribution can be efficiently approximated with Stochastic Gradient Variation Bayes (SGVB) \cite{kingma2013auto}, which uses a parametric recognition or {\it inference model} $q(z|y,x)$ to approximate the true posterior $p(z|y,x)$. The inference model learns a mapping from observations to latent space, providing generalization and thereby amortizing the cost of inference (compared to Markov chain Monte Carlo (MCMC) inference methods that needed computationally expensive iterative procedures to estimate the latent variables per datapoint).  

We can derive a variational lower bound $\mathcal{L}(y|x)$ on our data likelihood $p(y|x)$:
\begin{align}
\log p(y|x) &\geq \mathbb{E}_{z \sim q(z|x,y)} \bigg[ \log \frac{p(y,z|x)}{q(z|y,x)}  \bigg]  \nonumber \\
& = \mathbb{E}_{z \sim q(z|x,y)} [\log p(y|z,x)] - D_\text{KL}[q(z|x,y) \| p(z|x)] = \mathcal{L}(y|x;\theta,\phi) \label{elbo}
\end{align} 

where $\theta$ denotes the parameters in the generative network, $\phi$ denotes the parameters in the inference network and prior, and $D_\text{KL}$ denotes the Kullback-Leibler (KL) divergence. We can interpret the left-hand term of the last equation ($\log p(y|z,x)$) as the negative {\it reconstruction error}, which measures how well we reconstruct $y$ after sampling $z$. The right-hand term (KL divergence) ensures $q$ does not diverge too much from the prior $p$. This acts as a regularizer, and ensures that we can at test time (when we do not observe $y$) sample from $p(z|x)$ instead of $q(z|x,y)$. 

In practice, we slightly modify the objective in Eq. \ref{elbo}, where we use importance sampling \cite{burda2015importance} to obtain a tighter bound, and minimize a different distance function instead of the KL-divergence (namely $\alpha$-divergence with $\alpha$=0.5 \cite{depeweg2016learning}). Details are provided in Appendix \ref{vaetarget}.

\begin{figure}[t]
\centering
\includegraphics[width = 0.7\textwidth]{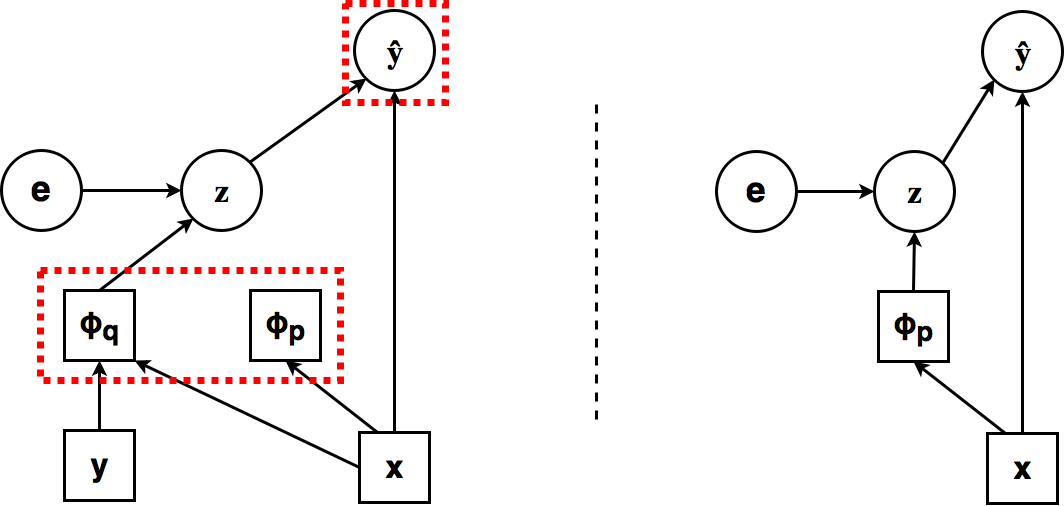}
\vspace{-0.2cm} 
\caption{Conditional Variational Auto-Encoder as a computational process. Squares are deterministic, circles are probabilistic nodes. {\bf Left}: Training procedure. During training, we sample $z$ according to $q(\cdot|x,y)$, where $q$ is parametrized by $\phi_q$. The training loss consists of two terms (indicated by the red dotted boxes): 1) the reconstruction loss $p(y|z,x)$, and 2) the KL-divergence between $q(z|x,y)$ and $p(z|x)$. The latter ensures that the posterior $q$ puts probability mass at the same points as the prior $p$, effectively acting as a regularizer in latent space. We compute $z$ with the reparametrization trick, where {\bf e} can be any appropriate noise distribution. {\bf Right}: Test procedure. At test time, we cut away the inference network $q$, and instead sample $z$ according to the prior $p(z|x)$. This allows us to make stochastic predictions for $y$.}
    \label{cvae2}
\end{figure}

\subsection{Reparametrization} \label{cont}
For this work we focus on variational methods that {\it reparametrize} the distribution of $q_\phi(z|y,x)$ to allow gradient-based training on a single computational graph. The trick works when we can write $z$ as a function $z=f_\phi(\epsilon,y,x)$, with $f_\phi(\cdot)$ a deterministic, differentiable function, and $\epsilon \sim p(\epsilon)$ a noise distribution with independent marginal. 

For a continuous variable $z$, the cardinal example is a location-scale transformation of a standard Gaussian distribution. If $q_\phi(z|y,x) = \mathcal{N}(z|\mu_\phi(y,x),\Sigma_\phi(y,x))$, then we can write

\begin{equation}
z = f_\phi(\epsilon,y,x) = \mu_\phi(y,x) + \Sigma_\phi(y,x) \cdot \epsilon,\quad  \text{with} \quad \epsilon \sim \mathcal{N}(0,1) \label{cont_repar}
\end{equation}

The gain is that we can now backpropagate through expectations of the random variable $z$ \cite{kingma2013auto}: 

\begin{equation}
\nabla_{\phi} \mathbb{E}_{z \sim q_\phi(z|x,y)}[\xi(z)] = \mathbb{E}_{\epsilon \sim \mathcal{N}(0,1)}\big[\nabla_{\phi} \xi(f_\phi(\epsilon,y,x))\big]
\end{equation}

for some function $\xi(\cdot)$ of $z$. The right-hand term can then be approximated with a Monte-Carlo estimate. This allows us to backpropagate through $z$ (see Fig. \ref{cvae2}, left), giving the vanilla conditional variational auto-encoder (CVAE) with Gaussian latent variables. The overall train and test procedure is summarized in Figure \ref{cvae2}. We now consider two methods to improve the capacity of the latent $z$ distribution, that both could improve learning multimodal outcomes.

\subsection{Discrete Latent Variables} \label{disc}
As we want to model multimodal outcomes, it seems natural to consider discrete latent variables. However, for the reparametrization trick to be applicable we require the function $f_\phi(\cdot)$ to be differentiable, which is not possible for a discrete variable. It turns out we can get good estimates by making a smooth approximation to the discrete loss \cite{jang2016categorical,maddison2016concrete}. 

Let $\omega_i$ be an ordered set of class probabilities of a discrete variable $z_i$\footnote{We use subscripts $z_i$ to index the elements of the vector random variable $z$, and double subscripts $z_{i,j}$ to index the categories within one discrete random variable.} with $n_i$ categories. We can draw samples from this distribution through the Gumbel-Max trick:

\begin{equation}
z_i = \mbox{one-hot} \bigg( \argmax_{j \in [1..n_i]} [ g_j + \log \omega_{i,j} ]  \bigg)
\end{equation} 

with $g_j$ i.i.d. draws from a Gumbel$(0,1)$ distribution\footnote{We can sample from a Gumbel(0,1) distribution by sampling $u \sim \mbox{Uniform}(0,1)$ and computing $g = -\log(-\log(u))$.}. Since $\argmax$ is not differentiable, we can make a softmax approximation to the above equation:

\begin{equation}
z_{i,j} = \frac{\exp((\log \omega_{i,j} + g_j)/\tau)}{\sum_{o=1}^{n_i} \exp((\log \omega_{i,o} + g_o)/\tau)} \quad \quad \mbox{for} \quad j = 1,..,n_i \label{eq20}
\end{equation}

which is known as the Gumbell-Softmax \cite{jang2016categorical} or Concrete \cite{maddison2016concrete} distribution. The softmax temperature $\tau \in (0,\infty)$ regulates the discreteness of the approximation: for $\tau \to 0$, the samples effectively become one-hot, while for $\tau \to \infty$, the samples become uniform over the class categories. The above specification allows us to use the reparametrization trick for discrete latent variables, as the noise distribution $g$ is now decoupled from the gradient path $\frac{\delta z}{\delta \omega}$. Note that Eq. \ref{eq20} is a type of reparametrization $f_\phi(\cdot)$ (as introduced in Sec. \ref{cont}, with $g$ the noise distribution and $\omega_\phi(x,y)$ the distribution parameters. In practice, we anneal $\tau$ from $>1$ to $0$ over the course of training. 

\subsection{Transformations of Continuous Variables (Flow)} \label{flow}
We already specified the reparametrization trick for spherical Gaussian latent variables (Eq. \ref{cont_repar}). As spherical Gaussians may be too restricting for multimodality, we can increase the capacity of the latent layer by using transformations of distributions for which we can track the density.

To obtain more expressive distributions for a continuous random variable $z \in \mathbb{R}^D$ with known density $q(z)$, we consider bijective smooth mappings $ h : \mathbb{R}^D \to \mathbb{R}^D $ with inverse $ h^{-1}$. We are interested in the distribution of the transformed variable $z' = h(z)$. As long as we are able to invert $h$, we can easily compute the density of the transformed variable $z'$:

\begin{equation}
q(z') = q(z) \bigg|\det (\frac{\delta h^{-1}(z')}{\delta z'})\bigg| = q(z) \bigg|\det (\frac{\delta h(z)}{\delta z})\bigg|^{-1}
\end{equation}

which is known as the {\it change-of-variable formula}. If we can specify our neural network to learn transformations which are easily invertible, we can construct more complicated distributions by repeatedly applying the above transformation (while being able to track the density). If we repeatedly apply a sequence of transformations $z^L = h^L \circ ... \circ h^1(z^0)$ for some random variable $z^0 \sim q^0(\cdot)$, then the density of the last variable $z^L$ can be computed as:

\begin{equation}
\log q^L(z^L) = \log q^0(z^0) - \sum_{l=1}^L \log \bigg| \det \frac{\delta z^l}{\delta z^{l-1}} \bigg|
\end{equation}

The problem with the above transformation is that, especially for high-dimensio\-nal domains, computing the determinant is computationally very expensive. An elegant solution appears from the observation that the determinant of a triangular matrix is simply the product of its diagonal terms \cite{dinh2016density} \cite{kingma2016improved}. Therefore, given a random variable $z$ of length $D$, we can specify the transformation $z'=h(z)$ as:

\begin{align} 
z'_{1:d} &= z_{1:d} \nonumber \\ 
z'_{d+1:D} &= t(z_{1:d}) + z'_{d+1:D} \odot \exp( s(z_{1:d})) \label{transformed}
\end{align}

The Jacobian of the this transformation is:

\begin{equation} 
\frac{\delta z'}{\delta z} = \begin{bmatrix} \mathbb{I}_d & 0 \\ \frac{\delta z'_{d+1:D}}{\delta z_{1:d}} & \mbox{diag}(\exp(s(z_{1:d}))) \end{bmatrix}
\end{equation}

The determinant of this matrix is easily computed as $ \exp\big[\sum_{i} s(z_{1:d})_i)\big]$. Note that the $t(\cdot)$ (translation) and $s(\cdot)$ (scale) function can be arbitrarily complex functions, for example deep, non-linear neural networks. In these transformations, we do not need to compute the determinant of $s(\cdot)$ or $t(\cdot)$ to track the density of the random variable $z'$. Moreover, it is trivial to invert the above transformation:

\begin{align} 
z_{1:d} &= z'_{1:d} \nonumber \\
z_{d+1:D} &= (z_{d+1:D} - t(z'_{1:d})) \odot \exp(- s(z'_{1:d}))
\end{align}

This allows us to use the change-of-variable formula of the previous section. We effectively perform an auto-regressive transformation on the $z$ variables. In practice, we repeatedly modify the order of the $z$ variables to have a different part of $z$ transformed in each layer. In Fig. \ref{cvae2}, we would apply these transformations to a sample from $q(z|x,y)$ before calculating the KL-divergence with $p(z|x)$.

\subsection{Enforcing Latent Variable Use} \label{freebits}
One of the challenges of training latent variable models is their tendency to overfit the prior early in training. Initially, the likelihood term $p(y|z,x)$ is relatively weak. Therefore, the learning signal is dominated by the KL-divergence, and stochastic optimization gets stuck in the undesirable equilibrium $q(z|y,x) \approx p(z|x)$.

To give a simple illustration, imagine $y$ is strictly bimodal given a fixed $x$, taking value $y_1$ or $y_2$ with $p(y_1|x)=0.3$ and $p(y_2|x)=0.7$. We fit a latent model with a single binary variable $z$ taking values $z_1$ or $z_2$. Clearly, we want our prior $p(z|x)$ to learn the distribution $\{p(z_1|x)=0.3,p(z_2|x)=0.7\}$ (assuming $z_1$ maps to $y_1$ and $z_2$ to $y_2$, which can of course be interchanged). However, the inference network $q(z|x,y)$ has access to additional information, as it knows which $y$ we need to reconstruct. Therefore, if we present a datapair $(x,y_1)$, then we want our latent distribution more like $\{q(z_1|x,y_1)=0.999,q(z_2|x,y_1)=0.001\}$, as this ensures we make a good draw and good reconstruction. However, for this datapoint this would incur a KL-penalty $D_{KL}(q(z|x,y_1)\|p(z|x)) \approx 1.20$. This illustrates how a good fitting VI model will necessarily incur some KL-cost.

A solution is to enforce each (set of) latent variables to encode a minimum amount of information \cite{kingma2016improved}, i.e. force $q(z|y,x)$ to at least have a KL-divergence of $\lambda$ from the prior $p(z|x)$. The modified objective becomes:
\vspace{-0.6cm}

\begin{multline}
 \mathcal{\tilde{L}}(y|x) = \mathbb{E}_{(x,y) \sim \mathcal{M}} \bigg[ \mathbb{E}_{z \sim q(z|y,x)} \big[ \log P(y|z,x) \big] \bigg] - \\ \sum_{j=1}^{D_z} \max\bigg(\lambda,\mathbb{E}_{(x,y) \sim \mathcal{M}} \big[ D_{KL}[q(z_j|x,y) \| p(z_j|x)] \big]\bigg) \label{kl_min}
\end{multline}

where $D_z$ is the dimensionality of the latent space $z$, and $\mathcal{M}$ denotes a mini-batch. Different solutions have been proposed, like KL annealing \cite{sonderby2016ladder}, but we empirically found them to be less effective.

\section{Results} \label{results}
We now test the different types of conditional variational inference, introduced in the previous section, on two tasks.  Evaluating generative model performance is not straightforward, as standard metrics like mean-squared error (MSE) are non-valid for multimodal outcome distributions. In this work, we evaluate {\bf i)} the log likelihood of a test set under the learned generative model (see Appendix \ref{evaluation}), and (if possible) {\bf ii)} we draw new data from the learned model and compute KL divergences or Hellinger distances with respect to the true data generating distribution. Training details and hyperparameters are described in Appendix \ref{hyper}.

\subsection{Toy Problem}

\begin{figure}[t] \centering
\includegraphics[width=1.0\textwidth]{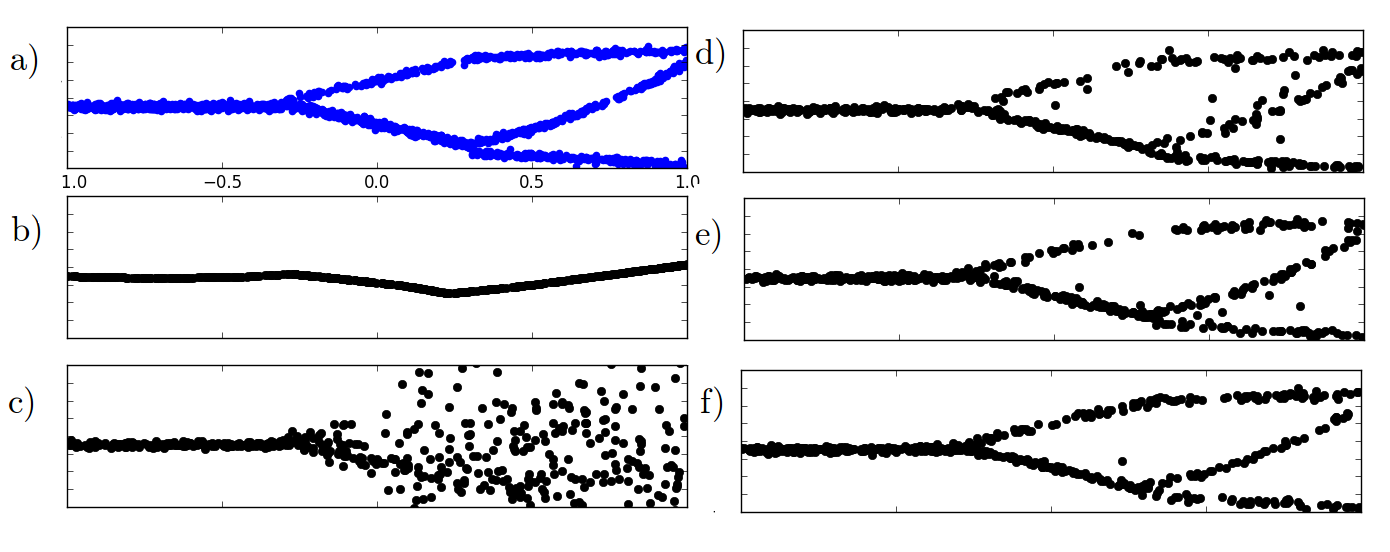}
\vspace{-0.9cm} 
\caption{Comparison of samples from the models produced by multi-layer perceptron (MLP) and variational inference (VI) networks after training for 30,000 mini-batches. a) Ground truth data. b) MLP (deterministic predictions). c) MLP with stochastic inputs. d) VI with spherical Gaussian. e) VI with spherical Gaussian and 5 layers of flow. f) VI with discrete latent variables. Numerical results are reported in Table \ref{toytable}.} \label{toyfigure}
\end{figure}

We generate a one-dimensional multimodal transition function by sampling $x \sim \mbox{Uniform}(-1,1)$ and sampling $y$ from a conditional Gaussian distribution $\mathcal{N}(\cdot|\mu=f(x),\sigma=0.1)$ according to:
\vspace{-0.1cm}
{\small
\begin{equation}
	p(y|x) = 
		\begin{cases}
		\mathcal{N}(2.5), & \mbox{if}\ x<-0.3 \\
		\rho_1 \mathcal{N}(4x) + \rho_2 \mathcal{N}(-4x), & \mbox{if}\ -0.3 \leq x<0.3 \\
		\rho_3 \mathcal{N}(5+\log(x+1)) + \rho_4 \mathcal{N}(-x + 0.2) + \rho_5 \mathcal{N}(5x^2), & \mbox{if}\ x \geq 0.3 \\
		\end{cases} \nonumber
\end{equation}
}
where $\rho_1=0.2$, $\rho_2=0.8$, $\rho_3=0.3$, $\rho_4=0.5$ and $\rho_5=0.2$. This generates the multimodal function shown in Figure \ref{toyfigure}a. We study this toy problem to visualize how different architectures will fit this simple data structure with conditional unimodal (left), bimodal (middle) and trimodal (right) structure (see Figure \ref{toyfigure}a left, middle and right parts). Figure \ref{toyfigure}b-f show the samples generated by different models after training on 30,000 mini-batch steps (See Appendix \ref{hyper} for details). Table \ref{toytable} displays the variational lower bound (VLB) and negative log-likelihood (NLL) on a test set. 

A feed-forward network trained on mean squared error deterministically predicts the conditional expectation (Figure \ref{toyfigure}b). For fair comparison, we also train a feed-forward network that does receive noise variables $\epsilon$ as input but without an inference network (Figure \ref{toyfigure}c). Theoretically this network could learn the same decoder distribution, but without the inference network the model does not converge. 

Figure \ref{toyfigure}d-f show the samples generated by different variational methods (spherical Gaussian {Sec. \ref{cont}), Gaussian with flow (Sec. \ref{flow}), and discrete latent variables (Sec. \ref{disc}), respectively. We see how these models are much better at fitting the true data distribution. Importantly, notice that the variational approach consistently predicts the deterministic part correctly (left part of the function). This is important, as the network is able to ignore the input noise when needed. Table \ref{toytable} indicates the discrete latent variable model fits this problem best.

\begin{table}[tb] \centering
\caption{Performance on toy domain. All results are averaged over 10 runs. VLB = Variational Lower Bound, NLL = Negative Log Likelihood on test dataset, MLP = Multi-Layer Perceptron, VAE = Variational Auto-Encoder.} \label{toytable} \vspace{-0.2cm}
\begin{tabular}{p{6cm}p{2cm}p{2cm}}
\hline
\noalign{\smallskip}
\bf Method & \bf VLB & \bf NLL  \\
\noalign{\smallskip} 
\hline
\noalign{\smallskip}
MLP (deterministic) & NA  & NA \\ 
MLP (with stochastic input) & NA  & 4.49 \\ 
VAE continuous (n=3, no flow) & 0.33  & -0.29  \\ 
VAE continuous (n=3, n$_{flow}$=5) & 0.32 & -0.33   \\ 
VAE discrete (n=3, k=3) & \bf 0.47 & \bf -0.48  \\ 
\hline
\end{tabular}
\end{table}

\subsection{Stochastic Gridworld} \label{stochgrid}
We now study a typical RL gridworld task with multimodal stochastic dynamics. The world is a 7x7 grid (see Figure \ref{predictionsfigure}) with some walls. The agent (green) starts in the bottom-left, can deterministically move in each cardinal direction, and needs to reach the top-right ($r=+10$). There are two ghosts, starting in locations as shown in Fig. \ref{onpolicyplot}, top-left. Ghost 1 (red) uniformly chooses one of the available directions. Ghost 2 (blue) has a bias to move to the left or right (40\% each), and moves vertically with small probability (10\%). Our interest here is to learn to predict this stochasticity from observed data. As state-space we use a vector of length 6 containing the 2D coordinates for the agent and both ghosts. Each element in the vector is treated as a categorical variable with 7 classes (for the 7x7 grid).\footnote{Note that, although this is a discrete MDP, it is not a trivial task to model multimodality here. Indeed, the outcome distribution {\it per state dimension} is categorical, but the joint distribution (generally) does not factorize over the dimensions. Therefore, we would already need a categorical with $7^6=117649$ outcome categories to learn this problem without conditional variational inference, and this would exponentially aggravate in larger state-spaces (e.g. images).}

\paragraph{\bf Uncorrelated Data.}
On of the core challenges of RL is the exploration problem, which can make the data we observe strongly correlated. For example, if the agent never explores the top-left region of the domain, we can not expect it to learn an accurate model there. To overcome this problem, we first study an idealized setting in which our dataset consists of the transitions of state-action combinations randomly sampled across state-space. 

\begin{figure}[t] \centering
\includegraphics[width=\textwidth]{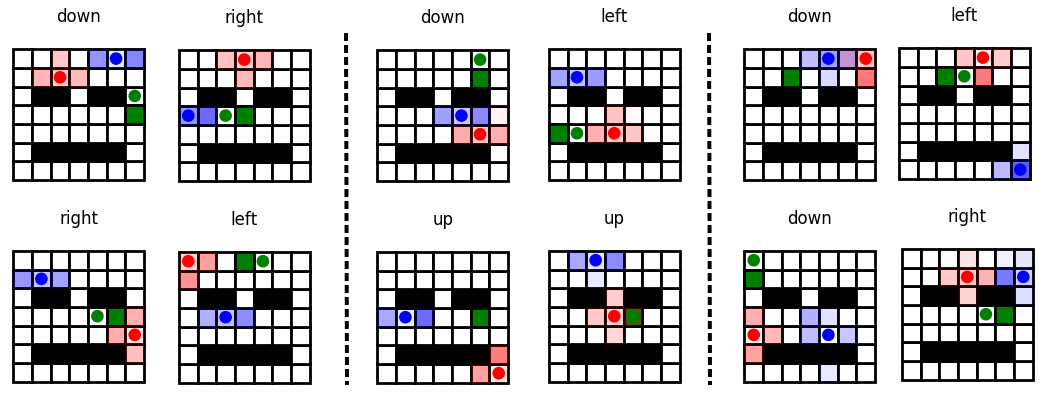}
\vspace{-0.7cm} 
\caption{Visual predictions on gridworld. Each sub-picture shows the agent (green), ghost 1 (red) and ghost 2 (blue) with current location as a circle, and predicted next location as a shaded box (color intensity corresponds to predicted probability). Black locations are walls, the text above each subplot indicates the action chosen by the agent. {\bf Left:} Continuous latent variables (n=8, no flow), {\bf Middle:} continuous latent variable (n=8, n$_{flow}$ =6), {\bf Right:} discrete latent variables (n=8,k=4). We observe stochastic predictions for the ghosts and deterministic predictions for the agent. Numerical comparison is provided in Table \ref{table2}.} \label{predictionsfigure}
\end{figure}

\begin{table}[t] \centering
\caption{Performance on gridworld predictions for different types of variational inference. For this table, $\hat{p}$ denotes the predicted distributions by the VAE model, while $p$ denotes the ground truth (which is known for this scenario). VLB = Variational Lower Bound, NLL = Negative Log Likelihood.} \label{table2}
\begin{tabular}{p{5.0cm}p{1cm}p{1cm}p{1.5cm}p{1.5cm}p{1.5cm}}
\hline\noalign{\smallskip}
\bf Method & \bf VLB & \bf NLL & \bf $D_{KL}(p\|\hat{p})$ & \bf $D_{Hel}(p\|\hat{p})$ & \bf $D_{KL}(\hat{p}\|p)$  \\ 
\noalign{\smallskip} 
\hline
\noalign{\smallskip}
VAE Continuous (n=8, no flow) & -2.53 & 2.52 & \bf 0.91 & \bf 0.48 & \bf 3.12  \\ 
VAE Continuous (n=8, n$_{flow}$=6) & -2.66 & 2.70 & 2.74 & 0.60 & 4.29  \\ 
VAE Discrete (n=8, k=4) & \bf -2.17 & \bf 2.20 & 1.26 & 0.61 & 4.75 \\ 
\hline
\end{tabular}
\end{table}

\paragraph{\bf On-policy Agent.}
\begin{figure}[ht] \centering
\includegraphics[width=1.0\textwidth]{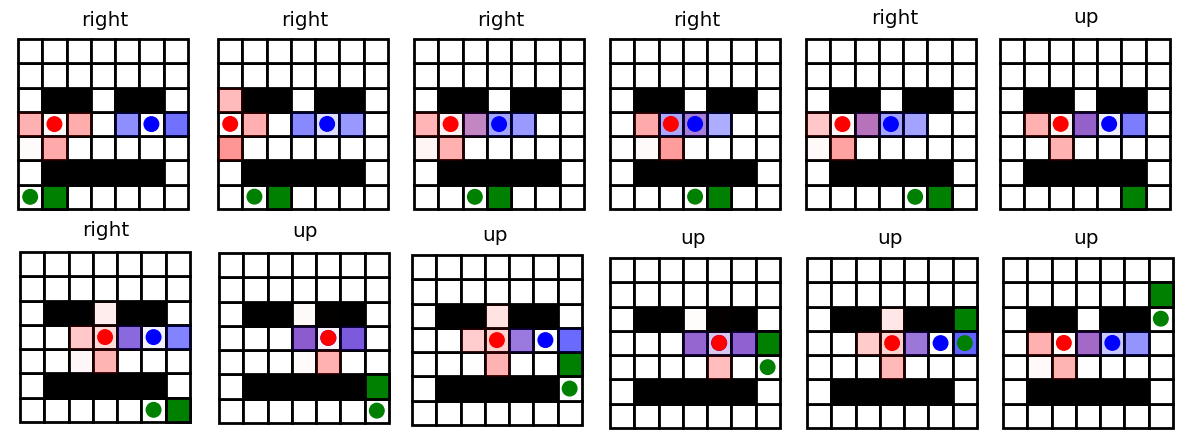}
\vspace{-0.4cm} 
\caption{{\it On-policy} predictions for RL agent (see Fig. \ref{predictionsfigure} for color explanation). The sub-plots progress row-wise along a roll-out in the learned transition model. Note that this is a true 12-step roll-out, i.e. each next plot is based on sampling a single prediction from the model (we do not observe any true next state along the way).} \label{onpolicyplot}
\end{figure}

The results on this task are shown in Table \ref{table2}. Compared to Table \ref{toytable} we do not show MLP results anymore for this task. We see the discrete latent variables again perform best on negative log-likelihood (NLL) evaluation. However, when we compare the learned distribution to the true distribution (which is available for this problem), we see the continuous latent variables without flow actually perform best. We see a conflict between both performance measures (the NLL indicates the discrete model performs best, while the distances with the true distribution point at the continuous latent model). In this case, visual comparison (Figure \ref{predictionsfigure}) does not show important differences across methods. We therefore conclude that the differences between methods are small for this problem, while the best performance measures for generative models remains an open questions in general (see \cite{goodfellow2016nips}). 

As a next step, we investigate to what extend an RL agent is capable of learning an accurate transition model {\it on-policy}, i.e. while observing correlated data. Note that the agent is still learning its policy in a model-free sense here (as a deep Q-network \cite{mnih2015human}), and we simply investigate to what extend the learned transition model is accurate after observing correlated data. Therefore, we evaluate the learned transition model while the agent is executing the policy. 

Figure \ref{onpolicyplot} shows the results of a roll-out in the learned model under a model-free policy. We see the agent first walking along the bottom corridor, and then moving up in the vertical corridor. Note that the agent consistently predicts its next state deterministically and correctly. In frame 6 it makes a wrong action decision, probably because we execute the behavioural policy with small $\epsilon$-greedy noise. The ghosts have multimodal, stochastic behaviour. The first ghost (red) moves uniformly in one of the available directions, which is captured by the red shades around the current ghost location. Note that the model consistently predicts the ghost to move at each step. The second ghost (blue) has a bias to primarily step to the left or right. We also note the difference in the predicted next state between red and blue ghost, matching their true dynamics. Altogether, the agent has learned to predict both the deterministic effects of its own actions as well as its stochastic environment, from on-policy, correlated data.

\vspace{-0.2cm}
\section{Related Work} \label{related}
\vspace{-0.2cm}
Variational inference in the conditional setting was previously studied by Sohn et al. \cite{sohn2015learning} and Walker et al. \cite{walker2016uncertain}. Compared to our work, these papers only use spherical Gaussian priors, and do not focus on reinforcement learning tasks. Our work focussed on VI with reparametrization gradients. There is a second line of research on latent variable models that uses score function gradients \cite{mnih2014neural}, which are also known as REINFORCE in the RL context. A benefit of reparametrization gradients is that they don't suffer from the high variance usually encountered with score function gradients (a problem also known in RL). 

The idea to apply flow to the latent layer originates from Rezende and Mohamed \cite{rezende2015variational}. The transformation in Eq. \ref{transformed} are related to the {\it affine coupling layers} of Dinh et al. \cite{dinh2016density}, but then applied to the latent layer of a CVAE, while Dnih et al. \cite{dinh2016density} use them directly from observation level without variational inference. Applying flow transformations at latent VI level was introduced by Kingma et al. \cite{kingma2016improved}, where the authors used fully autoregressive transformations (which are harder to implement compared to our transformations, but potentially have more representational capacity). 

To increase the expressivity of the latent approximation, we focussed on different types of latent variables, as well as (normalizing) flow. A third way to increase latent capacity is to factorize the distribution into several layers \cite{sonderby2016ladder}. However, activating deeper stochastic layers is not straightforward \cite{sonderby2016ladder}, requiring either batch normalization or weight normalization \cite{kingma2016improved}. We defer factorized inference networks to future work, especially in higher-dimensional tasks. 

The different deep generative models discussed in Section \ref{dgm} are not mutually exclusive. For example, the variational lossy auto-encoder (LVAE) \cite{chen2016variational} combines variational inference with PixelCNN-based decoders \cite{oord2016conditional}. Such architectures force high-level conceptual information into the latent level, while the decoder should capture fine-grained details. This could be beneficial to sparsify the latent layer and as such benefit RL planning as well. 

There is relatively little work on Bayesian Neural Networks for RL. Closest to ours is the work by Depeweg et al. \cite{depeweg2016learning}, who study VI to estimate both transition function stochasticity (as studied in this work) combined with uncertainty (due to limited data). Compared to their work, we use a parametric inference network which allows us to generalize in the inference part, while they perform VI per individual datapoint. Second, they only considered Gaussian latent variables, while we investigate discrete latent variables and normalizing flow as well. The results of Depeweg et al. \cite{depeweg2016learning} also show the ability to learn multimodal stochasticity, and additionally show the benefit of planning over the model. Watter et al. \cite{watter2015embed} also used variational auto-encoders in a control task, but only as a regularizer for learning representations, not to make stochastic predictions. Gal et al. \cite{gal2016improving} uses Bayesian neural networks, in the form of Bayesian dropout, to track uncertainty (due to limited data) in transition dynamics estimation.

Finally, there is also a line of RL research that uses the transition function target to speed-up model-free RL. This idea has been identified as RL with `auxiliary tasks' \cite{jaderberg2016reinforcement}. The gradients of the transition function predictions are denser compared to the sparse RL training signal, and used to speed-up training of deeper network layers shared between policy and transition network. However, this approach does not learn stochastic transitions (but could benefit from it, as it improves the learning signal), nor is it used for sample-based planning as in model-based RL.

\vspace{-0.2cm}
\section{Future Work} \label{future}
\vspace{-0.2cm}
One clear line of future work is to use these transition models to improve agent performance, by planning over the model with either a given or learned reward function. Depeweg et al. \cite{depeweg2016learning} already provided a study in this direction. Compared to their work, it would especially be interesting to apply more adaptive roll-outs in the model, like Monte Carlo Tree Search (MCTS). Moreover, it would be important to evaluate these methods in high-dimensional RL tasks, e.g. with convolutional neural networks on raw pixel data \cite{oh2015action}. Another extension is to use these models to improve exploration in stochastic domains (e.g. \cite{oh2015action,houthooft2016vime}).

An important second challenge, briefly mentioned in the Introduction and Section \ref{stochgrid}, is planning under uncertainty. RL initially provides correlated data from a limited part of state-space. When planning over this model, we should not extrapolate too much, nor trust our model too early with limited data. Planning under uncertainty was for example studied by Gal et al. \cite{gal2016improving} and Houthooft et al. \cite{houthooft2016vime}. Note that `uncertainty' (due to limited data) is fundamentally different from the `stochasticity' (true probabilistic nature of the domain) discussed in this paper.

A third challenge for transition dynamics estimation is memory (partial observability), when the current state does not provide all available information to make an prediction. Proposed solutions are recurrent neural networks (RNN)  or Neural Turing Machines (NTM), which have both been studied in the variational inference context (in \cite{chung2015recurrent} and \cite{gemici2017generative}, respectively). 

Combining stochasticity, uncertainty and memory in one function approximator would be an important integrating step in model-based RL.

\vspace{-0.2cm}
\section{Conclusion}
\vspace{-0.2cm}

This paper studied multimodal transition function estimation for RL agents, with a focus on variational inference with different types of latent variables. Our experiments show variational inference is a robust method to discriminate deterministic and stochastic elements of the transition function using function approximation, clearly improving over discriminative training. We verified results on a typical RL domain where tabular learning would be infeasible, showing the ability of these models to learn the multimodal transition dynamics online. We did not observe important distinction in performance between the different types of latent variables studied. Therefore, for the domain size studied in this work, it seems safe to use the standard spherical Gaussian conditional VAE. Our results are generally applicable in RL, and help solve a fundamental problem of many domains: the complex stochastic behaviour of its transition dynamics. Code to reproduce the results in this paper is publicly available at \url{www.github.com/tmoer/multimodal_varinf}. 

\bibliographystyle{splncs03}
{
\tiny
\bibliography{modelbased}
}
\clearpage

\appendix
\section{Variational Auto-Encoder (VAE) Training Objective} \label{vaetarget}
We can obtain a tighter bound on Equation \ref{elbo} by using importance sampling \cite{burda2015importance}. We sample $M$ values of $z$ per datapoint, and average over them inside the log. Otherwise, the model strongly penalizes for single samples that explain the objective poorly. Second, instead of the KL divergence we optimize Renyi $\alpha$-divergences \cite{li2016renyi}). We use $\alpha$=0.5 according to the results by Depeweg et al. \cite{depeweg2016learning}, which makes the divergence term become a function of the Hellinger distance \cite{li2016renyi}. The combined objective, known as the variational Renyi (VR) bound \cite{li2016renyi} is:


\begin{equation}
\mathcal{L}_{VR}(y|x) = \frac{1}{1-\alpha} \log \frac{1}{M} \sum_{m=1}^M \bigg[ \bigg( \frac{p(y,z^m|x)}{q(z^m|y,x)}\bigg)^{1-\alpha}  \bigg]
\end{equation}

with $z^m \sim q(\cdot|x,y)$.

\section{Test Set Negative Log-likelihood (NLL) for VAE} \label{evaluation}
We are interested in the likelihood $p(y|x)$ of a set of test data $\{x_i,y_i\}_{i=1}^N$. We therefore need to marginalize over $z$:
\vspace{-0.4cm}

\begin{equation}
p(y|x) = \mathbb{E}_{z \sim p(\cdot|x)} \bigg[ p(y|z,x) \bigg] 
\end{equation}
\vspace{-0.4cm}

One problem with this estimator is that we may need many empirical samples from $z$ to get an accurate estimate. As an alternative, we estimate the quantity through importance sampling, by sampling from $q(\cdot|x_i,y_i)$  instead of $p(\cdot|x_i)$:

\begin{equation}
p(y|x) = \mathbb{E}_{z \sim q(\cdot|x,y)} \bigg[ p(y|z,x) \frac{p(z|x)}{q(z|x,y)} \bigg]
\end{equation}
\vspace{-0.2cm}

The empirical estimate of the negative log likelihood (NLL), as reported in the results section, then becomes 
\vspace{-0.3cm}

\begin{equation} -\log p(y|x) = -\frac{1}{N} \sum_{i=1}^N  \log \Big[\frac{1}{M} \sum_{m=1}^M p(y_i|z_i^m,x_i) \frac{p(z_i^m|x_i)}{q(z_i^m|x_i,y_i)} \Big]
\end{equation}

with $z_i^m \sim q(\cdot|x_i,y_i)$.

\section{Training Details} \label{hyper}
For all experiments we follow standard train, validation and test set set-up. For all domains, we train the VAE target on $k=3$ importance samples with Renyi-$\alpha$ divergence for $\alpha=0.5$ (see appendix \ref{vaetarget}). This gave us slightly better results compared to the `default' settings of $k=1$ and $\alpha = 1.0$. All models are trained in Tensorflow using Adam optimizer. 
\newline 

{\bf Toy Domain}: We draw a training set of size 2000, and independent validation and test sets of size 500 and 2000, respectively. The decoder distribution is Gaussian, where we also learn its standard deviation. We train for 30000 batches with batch size 64, with a learning rate linearly annealed from 0.005 to 0.0005 over 90\% of training steps. The minimal KL penalty per dimension $\lambda$ (Eq. \ref{kl_min}) is fixed at 0.07. 

The generative network has three layers with 50 units per layer and Relu non-linearities. The inference network has two layers with 30 units per layer and Relu non-linearities. For the discrete latent variables, we anneal the temperature from 2.0 to 0.001 over 70\% of training steps. 
\newline 

{\bf Gridworld}: For the first task, we repeatedly draw training data by sampling a new state-action combination uniformly across state-space, and sampling a single transition.  Optimal model performance is based on a VAE performance on a validation and test set of size 750 and 1500 respectively. The decoder distribution is discrete taking values in 7 categories. We train on mini-batches of size 32 for 75000 iterations, with a learning rate linearly annealed from 0.0005 to 0.0001 over 70\% of training steps. The generative network has three layers with 250 units per layer and Relu non-linearities. The inference network has two layers with 100 units per layer and Relu non-linearities. The minimal KL penalty per dimension $\lambda$ (Eq. \ref{kl_min}) is fixed at 0.07. 

For the on-policy evaluation, the RL policy is trained as a deep Q-network \cite{mnih2015human} with target network and no experience replay. The state-action value network has three layers of 50 units and Relu activations. Given a mini-batch $\mathcal{M}$ of roll-out data under the current policy, the network is trained on the 1 step Q-learning objective:

\begin{equation}
L_{RL}(\eta) = \mathbb{E}_{(s,a,r,s')  \sim \mathcal{M}} \bigg[ \bigg( r + \gamma \max_{a'} Q(s',a';\eta^-) - Q(s,a;\eta)  \bigg)^2 \bigg]
\end{equation}

where $s,a,r$ denote state, action and reward, $Q(s,a)$ is the expected discounted return (discount parameter $\gamma = 0.99$) from state $s$ and action $a$ under the current policy, $\eta$ are the parameters in the value function network, and $\eta^-$ the parameters in the target network (which are fixed in the above loss, and only updated every 500 steps). During learning, we follow an $\epsilon$-greedy policy with $\epsilon$ linearly decayed from 1.0 to 0.10 over 60\% of training steps. 
\end{document}